\pdfoutput=1
\documentclass[11pt]{article}

\usepackage[]{ACL2023}

\usepackage{times}
\usepackage{latexsym}

\usepackage[T1]{fontenc}
\usepackage[utf8]{inputenc}

\usepackage{microtype}

\usepackage[figuresright]{rotating}
\usepackage{amsfonts,amssymb,amsmath,amsthm,amsopn,mathrsfs,mathtools,wasysym}	% math related
\usepackage{hhline,booktabs,colortbl,multirow,tabularx,diagbox,tablefootnote,threeparttable} % table related
\usepackage{graphicx}
\usepackage{algorithm}
\usepackage{algorithmic}
\usepackage{subfig}
\usepackage{prettyref}
\usepackage{adjustbox} 
\usepackage{rotate}
\usepackage{enumitem}
\usepackage{hyperref}
\usepackage{pifont}
\usepackage{listings}

\newcommand{\xmark}{\ding{55}}%

\def\our{\texttt{PyABSA}}

\usepackage{xcolor}

\definecolor{mycyan}{gray}{.7}

\newcommand{\pref}{\prettyref}

\newrefformat{fig}{Figure~\ref{#1}}
\newrefformat{tab}{Table~\ref{#1}}
\newrefformat{sec}{Section~\ref{#1}}
\newrefformat{app}{Appendix~\ref{#1}}
\newrefformat{alg}{Algorithm~\ref{#1}}
\newrefformat{property}{Property~\ref{#1}}
\newrefformat{theorem}{Theorem~\ref{#1}}
\newrefformat{corollary}{Corollary~\ref{#1}}
\newrefformat{proposition}{Proposition~\ref{#1}}
\newrefformat{def}{Definition~\ref{#1}}
\newrefformat{eq}{equation~(\ref{#1})}

\usepackage{newfloat}
\DeclareFloatingEnvironment[
    fileext=los,
    listname={List of Snippets},
    name=Snippet,
    placement=htbp,
]{snippet}

\title{\our: A Modularized Framework for Reproducible Aspect-based Sentiment Analysis}

\author{
	Heng Yang$^1$, Chen Zhang$^2$, Ke Li$^1$ \\ 
	$^1$Department of Computer Science, University of Exeter, EX4 4QF, Exeter, UK \\
        $^2$School of Computer Science, Beijing Institute of Technology, Beijing, China \\
	\texttt{\{hy345,k.li\}@exeter.ac.uk}, \texttt{czhang@bit.edu.cn} \\
}

\begin{document}
\maketitle
\begin{abstract}
The advancement of aspect-based sentiment analysis (ABSA) has urged the lack of a user-friendly framework that can largely lower the difficulty of reproducing state-of-the-art ABSA performance, especially for beginners. To meet the demand, we present \our, a modularized framework built on PyTorch for reproducible ABSA. To facilitate ABSA research, \our\ supports several ABSA subtasks, including aspect term extraction, aspect sentiment classification, and end-to-end aspect-based sentiment analysis. Concretely, \our\ integrates 29 models and 26 datasets. With just a few lines of code, the result of a model on a specific dataset can be reproduced. With a modularized design, \our\ can also be flexiblely extended to considered models, datasets, and other related tasks. Besides, \our\ highlights its data augmentation and annotation features, which significantly address data scarity. All are welcome to have a try at \url{https://github.com/yangheng95/PyABSA}.

\end{abstract}

% \textcolor{red}{[TODO: 1) fix code snippet; 2) fix framework fig; 3) fix model table; 4) more snippets? 5) some missing references.]}

\section{Introduction}
Aspect-based sentiment analysis (ABSA)~\cite{PontikiGPPAM14,PontikiGPMA15,PontikiGPAMAAZQ16} has made remarkable strides in recent years, particularly in the subtasks of aspect term extraction (ATE)~\cite{YinWDXZZ16,WangPDX16,LiL17,WangPDX17,LiBLLY18,XuLSY18,MaLWXW19,Yang2019,YangLQSS20}, aspect sentiment classification (ASC) ~\cite{MaLZW17,ZhangLS19,HuangC19a,PhanO20,ZhaoHW20,LiCFMWH20,DaiYSLQ21,TianCS21,WangSLZC21}, and end-to-end aspect-based sentiment analysis (E2EABSA)~\cite{YangZYSX21}. In an example sentence that ``I love the \textit{pizza} at this restaurant, but the \textit{service} is terrible.'', there are two aspects "\textit{pizza}" and ``\textit{service}''. towards which the sentiments are positive and negative, respectively. Here, ATE aims to extract the two aspects, ASC aims to detect the corresponding sentiments given the aspects, and E2EABSA\footnote{There are aliases for ASC and E2EABSA in some research, i.e., APC and ATEPC.} aims to achieve the extraction and detection as one. 

While an enormous number of models have been proposed in ABSA, however, they typically have distinguished architectures (e.g., LSTM, GCN, BERT) and optimizations (e.g., data pre-processing, evaluation metric), making it hard to reproduce their reported results even if their code is released. To address this issue and promote a fair comparison, we introduce \our, a modularized framework built on PyTorch for reproducible ABSA. We provide a demonstration video\footnote{The video can be accesses at: \url{https://www.youtube.com/watch?v=Od7t6CuCo6M}} to show the basic usages of \our.

\our~enables easy-to-use model training, evaluation, and inference on aforementioned ABSA subtasks with 29 models and 26 datasets supported. \our\ allows beginners to reproduce the result of a model on a specific dataset with just a few lines of code.
%\footnote{We provide an inference service demo on \url{https://huggingface.co/spaces/yangheng/Multilingual-Aspect-Based-Sentiment-Analysis}} %We present a ATESC demo based on \our\ in \pref{fig:demo}. 
In addition to using \our\ to reproduce results, we have also released a range of trained checkpoints, which can be accessed through the Transformers Model Hub\footnote{The \href{https://huggingface.co/spaces/yangheng/Multilingual-Aspect-Based-Sentiment-Analysis/tree/main/checkpoint}{\underline{Model Hub}} of \our\ is powered by Huggingface Space.} for users who need exact reproducibility. 

Moreover, \our\ is a framework with a modularized organization. Technically, \our\ has five major modules: template class, configuration manager, dataset manager, metric visualizer, checkpoint manager. Thus it is flexible to extend provided templates to considered models, datasets and other related tasks with minor modifications.

% \begin{figure*}[ht]
% 	\centering
% 	\includegraphics[width=\textwidth]{figs/demo1.pdf}
% 	\caption{An illustration of the capabilities of our multilingual aspect term extraction and sentiment classification demo. This demo has been designed to be user-friendly, allowing anyone to extract aspect terms and classify sentiments from multilingual user inputs with ease. It has already provided assistance to many users as they develop their own projects.}
% 	\label{fig:demo}
% \end{figure*}

%Thus, in an ABSA task, the goal would be to extract these two aspects, ``pizza'' and ``service'', and to classify the sentiment towards each aspect as positive or negative. The result of ABSA on this sentence would be: Users can access an available ABSA inference model from the hub-hosted model checkpoint through a single API without any training~\footnote{The examples can be found in the documentation at: \url{https://pyabsa.readthedocs.io.}}. \our\ includes over $40$ models for performing ATE and ASC subtasks, some of which have achieved state-of-the-art performance according to \href{https://paperswithcode.com/sota/aspect-based-sentiment-analysis-on-semeval}{Paperswithcode}\footnote{Our Fast-LSA model and LCF-ATEPC model achieve state-of-the-art performance on ASC subtask.}. It also offers multi-task-based ATESC models that perform both ATE and ASC subtasks simultaneously. 

It is widely recognized that ABSA models suffers from the shortage of data and the absence of datasets in specific domains. %\our\ currently offers up to $23$ ABSA datasets\footnote{We release the datasets at \url{https://github.com/yangheng95/ABSADatasets}} in $8$ languages, covering a wide range of domains. To the best of our knowledge, \our\ offers the largest number of open-source ABSA datasets compared to other open-source projects. 
Utilizing an ABSA-oriented data augmentor, we are able to provide up to $200K+$ additional examples per dataset. The augmented datasets can improve the accuracy of models by $1-3\%$. To encourage the community to contribute custom datasets, we provide an data annotation interface.

It is noteworthy that there are other existing projects partly achieving similar goals with \our. We should mark that the advantages of \our\ over these projects are in the following aspects.
\begin{itemize}[leftmargin=*,nolistsep,noitemsep]
    \item \our\ democratizes reproducible ABSA research by supporting a larger array of models and datasets among mainly concerned ABSA subtasks.
    \item \our\ is a modularized framework that is flexible to be extended to considered models, datasets, and other related tasks thanks to its organization.
    \item \our\ additionally offers data augmentation and data annotation features to address the data scarity in ABSA. 
\end{itemize}

\section{Supported Tasks}

\begin{table}[hbtp]
  \centering
  \caption{The prevalent models provided by \our. ATE and E2EABSA share similar models. Note that the models based on \texttt{BERT} can be adapted to other pre-trained language models from HuggingFace Transformers.}
  \resizebox{\linewidth}{!}{
    \begin{tabular}{|c|c|c|c|c|}
    \hline
    \textbf{Model} & \textbf{Task}  & \textbf{Reference} & \textbf{GloVe} & \textbf{BERT} \\
    \hline
    \hline
    \texttt{AOA}   & \multirow{24}[24]{*}{\begin{sideways}ASC\end{sideways}}  & \citet{HuangOC18} & \checkmark & \checkmark \\
    \cline{1-1}    \texttt{ASGCN}         &       & \citet{ZhangLS19} & \checkmark & \checkmark \\
    \cline{1-1}    \texttt{ATAE-LSTM}     &       & \citet{WangHZZ16} & \checkmark & \checkmark \\
    \cline{1-1}    \texttt{Cabasc}        &       & \citet{LiuZZHW18} & \checkmark & \checkmark \\
    \cline{1-1}    \texttt{IAN}           &       & \citet{MaLZW17}   & \checkmark & \checkmark \\
    \cline{1-1}    \texttt{LSTM-ASC}      &       & \citet{HochreiterS97} & \checkmark & \checkmark \\
    \cline{1-1}    \texttt{MemNet}        &       & \citet{TangQL16}  & \checkmark & \checkmark \\
    \cline{1-1}    \texttt{MGAN}          &       & \citet{FanFZ18}   & \checkmark & \checkmark \\
    \cline{1-1}    \texttt{RAM}           &       & \citet{ChenSBY17} & \checkmark & \checkmark \\
    \cline{1-1}    \texttt{TC-LSTM}       &       & \citet{TangQFL16} & \checkmark & \checkmark \\
    \cline{1-1}    \texttt{TD-LSTM}       &       & \citet{TangQFL16} & \checkmark & \checkmark \\
    \cline{1-1}    \texttt{TNet-LF}       &       & \citet{LamLSB18}  & \checkmark & \checkmark \\
    \cline{1-1}    \texttt{BERT-ASC}      &       & \citet{DevlinCLT19} & \xmark & \checkmark \\
    \cline{1-1}    \texttt{BERT-SPC}      &       & \citet{DevlinCLT19} & \xmark & \checkmark \\
    \cline{1-1}    \texttt{DLCF-DCA}      &       & \citet{XuZYCCL22} & \xmark & \checkmark \\
    \cline{1-1}    \texttt{DLCFS-DCA}     &       & \citet{XuZYCCL22} & \xmark & \checkmark \\
    \cline{1-1}    \texttt{Fast-LCF-ASC}  &       & \citet{ZengYXZH2019} & \xmark & \checkmark \\
    \cline{1-1}    \texttt{Fast-LCFS-ASC} &       & \citet{ZengYXZH2019} & \xmark & \checkmark \\
    \cline{1-1}    \texttt{LCA-BERT}      &       & \citet{YangZ2020} & \xmark & \checkmark \\
    \cline{1-1}    \texttt{LCF-BERT}      &       & \citet{ZengYXZH2019} & \xmark & \checkmark \\
    \cline{1-1}    \texttt{LCFS-BERT}     &       & \citet{ZengYXZH2019} & \xmark & \checkmark \\
    \cline{1-1}    \texttt{Fast-LSA-T}    &       & \citet{YangZXW2021} & \xmark & \checkmark \\
    \cline{1-1}    \texttt{Fast-LSA-S}    &       & \citet{YangZXW2021} & \xmark & \checkmark \\
    \cline{1-1}    \texttt{Fast-LSA-P}    &       & \citet{YangZXW2021} & \xmark & \checkmark \\
    \hline
    \hline
    \texttt{BERT-ATESC}     & \multirow{5}[5]{*}{\begin{sideways}ATE~/~E2E\end{sideways}}     & \citet{DevlinCLT19}& \xmark & \checkmark \\
    \cline{1-1}    \texttt{Fast-LCF-ATESC}  &       & \citet{YangZYSX21} & \xmark & \checkmark \\
    \cline{1-1}    \texttt{Fast-LCFS-ATESC} &       & \citet{YangZYSX21} & \xmark & \checkmark \\
    \cline{1-1}    \texttt{LCF-ATESC}       &       & \citet{YangZYSX21} & \xmark & \checkmark \\
    \cline{1-1}    \texttt{LCFS-ATESC}      &       & \citet{YangZYSX21} & \xmark & \checkmark \\
    \hline
    \end{tabular}%
    }
  \label{tab:models}%
\end{table}%

We primarily support three subtasks in ABSA, namely ATE, ASC, and E2EABSA. Each subtask contains its own models and datasets, which adds up to 29 models and 26 datasets in total.

\subsection{Models \& Datasets}
% This section introduces the built-in models in \our, which include state-of-the-art \texttt{LSA} model for aspect-based sentiment classification and \texttt{LCF-ATESC} models for aspect term extraction. Additionally, we have incorporated some popular ASC models from \texttt{ABSA-PyTorch} to provide users with a wider range of options. 

The core difficulty in unifying different models into a framework is that distinguished architectures and optimizations being used. We strive to bridge the gap in \our, which has to our best knowledge the largest model pool covering attention-based, graph-based, and BERT-based models, etc. The supported models are listed in \pref{tab:models}.

\our\ also gathers a wide variety of datasets across various domains and languages, including laptops, restaurants, MOOCs, Twitter, and others. As far as we know, \our\ maintains the largest ever number of ABSA datasets, which can be viewed in \pref{tab:datasets}. 

With just a few lines of code, researchers and users can invoke these builtin models and datasets for their own purposes. An example training pipeline of ASC is given in Snippet~\ref{snp:train}.

\begin{snippet}[ht]
\caption{The code snippet of an ASC training pipeline.}
\label{snp:train}
\begin{lstlisting}[language=Python,basicstyle=\ttfamily\tiny,frame=shadowbox,keywordstyle=\color{red},breaklines=True]
from pyabsa import AspectSentimentClassification as ASC

config = ASC.ASCConfigManager.get_asc_config_multilingual()
config.model = ASC.ASCModelList.FAST_LSA_T_V2

datasets_path = ASC.ABSADatasetList.Multilingual  
sent_classifier = Trainer(config=config,
                          dataset=datasets_path,
                          checkpoint_save_mode=1,  # save state_dict instead of model
                          auto_device=True,  # auto-select cuda device
                          load_aug=True,  # training using augmentation data
                          ).load_trained_model()

\end{lstlisting}
\end{snippet}

\begin{table*}[htbp]
  \centering
  \caption{
  A list of datasets in various languages presented in \our, where the datasets marked with $^\dagger$ are used for adversarial research. The increased number of examples in the training set have been generated using our own ABSA automatic augmentation tool.
  }
    \resizebox{.9\linewidth}{!}{
    \begin{tabular}{|c|c|c|c|c|c|c|}
    \hline
    \multirow{1}[4]{*}{\textbf{Dataset}} & \multirow{1}[4]{*}{\textbf{Language}} & \multicolumn{3}{c|}{\textbf{\# of Examples}} & \textbf{\# of Augmented Examples} & \multirow{1}[4]{*}{\textbf{Source}} \\
\cline{3-6}         & & Training Set & Validation Set & Testing Set & Training Set &  \\
    \hline
    \hline
    \texttt{Laptop14} & English & 2328  & 0     & 638   & 13325 &  \href{https://alt.qcri.org/semeval2014/task4/index.php?id=data-and-tools}{SemEval 2014}\\
    \cline{1-1}    \texttt{Restaurant14} & English & 3604  & 0     & 1120  & 19832 &  \href{https://alt.qcri.org/semeval2014/task4/index.php?id=data-and-tools}{SemEval 2014}\\
    \cline{1-1}    \texttt{Restaurant15} & English & 1200  & 0     & 539   & 7311  &  \href{https://alt.qcri.org/semeval2015/task12/index.php?id=data-and-tools}{SemEval 2015}\\
    \cline{1-1}    \texttt{Restaurant16} & English & 1744  & 0     & 614   & 10372 &  \href{https://alt.qcri.org/semeval2016/task5/index.php?id=data-and-tools}{SemEval 2016}\\
    \cline{1-1}    \texttt{Twitter} & English & 5880  & 0     & 654   & 35227 &  \citet{DongWTTZX14}\\
    \cline{1-1}    \texttt{MAMS}  & English & 11181 & 1332  & 1336  & 62665 &  \citet{JiangCXAY19}\\
    \cline{1-1}    \texttt{Television} & English & 3647  & 0     & 915   & 25676 &  \citet{MukherjeeSCMDG21}\\
    \cline{1-1}    \texttt{T-shirt} & English & 1834  & 0     & 465   & 15086 &  \citet{MukherjeeSCMDG21}\\
    \cline{1-1}    \texttt{Yelp}  & English & 808   & 0     & 245   & 2547  & \href{https://github.com/yangheng95/ABSADatasets}{WeiLi9811@GitHub}\\
    \cline{1-1}    \texttt{Phone} & Chinese & 1740  & 0     & 647   & 0     &  \citet{PengMLC18}\\
    \cline{1-1}    \texttt{Car}   & Chinese & 862   & 0     & 284   & 0     &  \citet{PengMLC18}\\
    \cline{1-1}    \texttt{Notebook} & Chinese & 464   & 0     & 154   & 0     &  \citet{PengMLC18}\\
    \cline{1-1}    \texttt{Camera} & Chinese & 1500  & 0     & 571   & 0     &  \citet{PengMLC18}\\
    \cline{1-1}    \texttt{MOOC}  & Chinese & 1583  & 0     & 396   & 0     &  \href{https://github.com/jmc-123/ABSADatasets}{jmc-123@GitHub}\\
    \cline{1-1}    \texttt{Shampoo} & Chinese & 6810  & 0     & 915   & 0     &  \href{https://github.com/brightgems/ABSADatasets}{brightgems@GitHub}\\
    \cline{1-1}    \texttt{MOOC-En} &  English & 1492  & 0     & 459   & 10562 &  \href{https://github.com/aparnavalli/ABSADatasets}{aparnavalli@GitHub}\\
    \cline{1-1}    \texttt{Arabic} & Arabic & 9620  & 0     & 2372  & 0     &  \href{https://alt.qcri.org/semeval2016/task5/index.php?id=data-and-tools}{SemEval 2016}\\
    \cline{1-1}    \texttt{Dutch} & Dutch & 1283  & 0     & 394   & 0     &  \href{https://alt.qcri.org/semeval2016/task5/index.php?id=data-and-tools}{SemEval 2016}\\
    \cline{1-1}    \texttt{Spanish} & Spanish & 1928  & 0     & 731   & 0     &  \href{https://alt.qcri.org/semeval2016/task5/index.php?id=data-and-tools}{SemEval 2016}\\
    \cline{1-1}    \texttt{Turkish} & Turkish & 1385  & 0     & 146   & 0     &  \href{https://alt.qcri.org/semeval2016/task5/index.php?id=data-and-tools}{SemEval 2016}\\
    \cline{1-1}    \texttt{Russian} & Russian & 3157  & 0     & 969   & 0     &  \href{https://alt.qcri.org/semeval2016/task5/index.php?id=data-and-tools}{SemEval 2016}\\
    \cline{1-1}    \texttt{French} & French & 1769  & 0     & 718   & 0     &  \href{https://alt.qcri.org/semeval2016/task5/index.php?id=data-and-tools}{SemEval 2016}\\
    \cline{1-1}    \texttt{ARTS-Laptop14$^\dagger$} & English & 2328  & 638   & 1877  & 13325 &  \citet{XingJJWZH20}\\
    \cline{1-1}    \texttt{ARTS-Restaurant14$^\dagger$} & English & 3604  & 1120  & 3448  & 19832 &  \citet{XingJJWZH20}\\
    \cline{1-1}    \texttt{Kaggle$^\dagger$} & English & 3376  & 0   & 866  & 0 &  \href{https://www.kaggle.com/datasets/cf7394cb629b099cf94f3c3ba87e1d37da7bfb173926206247cd651db7a8da07}{Khandeka@Kaggle}\\
    \cline{1-1}    \texttt{Chinese-Restaurant$^\dagger$} & Chinese & 26119  & 3638   & 7508  & 0 &  \citet{ZhangRMW0022}\\
    \hline
    \end{tabular}%
    }
  \label{tab:datasets}%
\end{table*}%

\subsection{Reproduction}

We as well present a preliminary performance overview of the models over the datasets provided in \our. The results, which are based on ten epochs of training using the configurations for reproduction, can be found in Appendix~\ref{app:eval}. The standard deviations of the results are also attached in parentheses. We used the pile of all datasets from \our~as the multilingual one. Please note that "-" in the results table means that the graph-based models are not applicable for those specific datasets. The checkpoints of these models are also offered for exact reproducibility. 
% An E2E ABSA example inference pipeline is given in \pref{fig:output} and Snippet~\ref{snp:infer}.
An E2E ABSA example inference pipeline is given in Snippet~\ref{snp:infer}.

\begin{snippet}[ht]
\caption{The code snippet of an E2EABSA inference pipeline.}
\label{snp:infer}
\begin{lstlisting}[language=Python,basicstyle=\ttfamily\tiny,frame=shadowbox,keywordstyle=\color{red},breaklines=True]
from pyabsa import AspectTermExtraction as ATE

aspect_extractor = ATE.AspectExtractor(
    "multilingual",
    data_num=100,
)
# simple inference
examples = [
    "But the staff was so nice to us .",
    "But the staff was so horrible to us .",
]
result = aspect_extractor.predict(
        example=examples,   
        print_result=True,  # print results in console
        ignore_error=True,  # ignore an invalid input
        eval_batch_size=32, # set batch size
    )
# batch inference
atepc_result = aspect_extractor.batch_predict(
    inference_source,  
    save_result=False,
    print_result=True, 
    pred_sentiment=True, 
    eval_batch_size=32,
)

\end{lstlisting}
\end{snippet}

% \begin{figure*}[ht]
% 	\centering
%     \caption{The console output of E2EABSA using \our.}
% 	\includegraphics[width=0.8\linewidth]{figs/output.png}
% 	\label{fig:output}
% \end{figure*}

\section{Modularized Framework}

\begin{figure*}[ht]
	\centering
	\includegraphics[width=\linewidth]{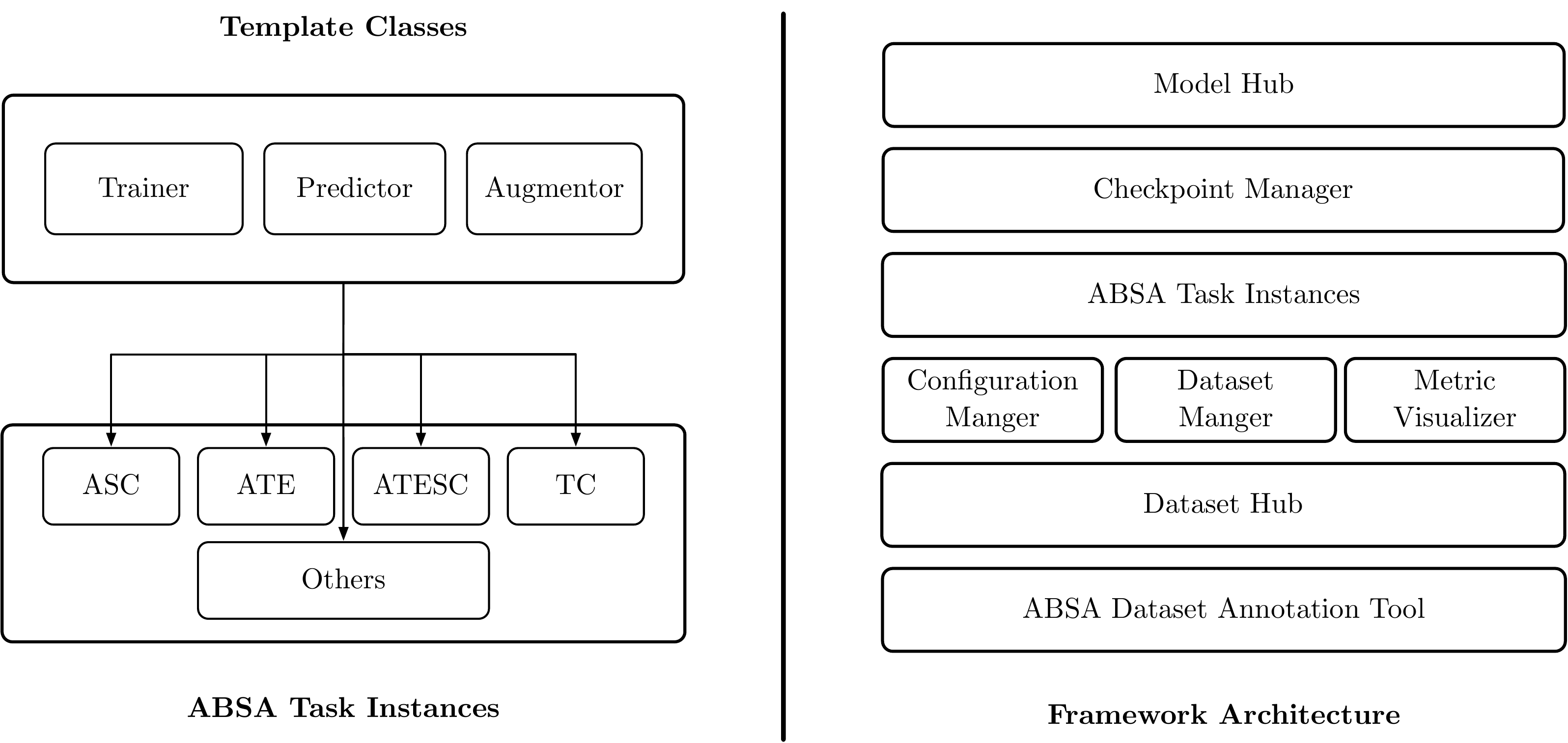}
	\caption{
 The left half of the diagram introduces the template classes provided in \our. Typically, each ABSA subtask has 5 template classes that need to be instantiated, except for the augmenter which is optional.The right side of the diagram shows the main framework of \our. The lowest level is the data annotation, which is suitable for creating custom datasets and the created datasets can be shared to the dataset hub. The three modules in the middle are the generic modules, which are suitable for training based on new datasets or models. The checkpoint manager is used to connect to the model hub and is responsible for uploading and downloading models and instantiating inference models.
 }
\label{fig:framework}
\end{figure*}

The main design of \our\ is shown in \pref{fig:framework}, which includes five necessary modules. We start by exploring task instances, which are abstracted as template classes. Afterwards, we dive into other modules (i.e., configuration manager, dataset manager, metric visualizer, checkpoint manager), elaborating their roles in getting \our\ modularized.

\subsection{Template Classes}

\our\ streamlines the process of developing models for ABSA subtasks, with a range of templates (refer to the five template classes in \pref{fig:framework}) that simplify the implementation of models and ease the customization of data.

We follow a software engineering design with common templates and interfaces, allowing users to defining models with model utilities, processing data with data utilities, training models with trainer, and inferring models with predictors. These can be all achieved simply by inheriting the templates without manipulating the common modules. The inherited modules come with a uniform interface for all task-agnostic features.

\subsection{Configuration Manager}
Configuration manager handles environment configurations, model configurations, and hyperparameter settings. It extends the Python Namespace object for improving user-friendliness. Additionally, The configuration manager possesses a configuration checker to make sure that incorrect configurations do not pass necessary sanity checks, helping users keep in track of their training settings. 

% The configuration manager also integrates with \our's model hub, saving and loading the manager object in sync with the model checkpoint. 
% Finally, \our\ provides access to configuration details during both the training and inference phases, making it easy for users to refer to these details at any time.

\subsection{Dataset Manager}
Dataset manager enables users to manage a wide range of builtin and custom datasets. Each dataset is assigned with unique ID and name for management, and the dataset items are designed as nest objects to enhance flexibility. This design makes it simple to combine datasets for ensemble learning and multilingual ABSA tasks. Moreover, the dataset manager also takes care of seamlessly connect to the ABSA dataset hub, automatically downloading and managing the integrated datasets.

 % In particular, we include both automatic and manual dataset annotation methods in ABSA Dataset Utils, enabling users to use \our\  for custom datasets compared to existing work. This is a really effective way to annotate custom datasets. In the case of community-contributed Yelp datasets, the built-in data augmentation method can improve performance by 2\% to 4\% (refer to \pref{sec:annotation}). Before feeding the data into the model, the dataset manager delivers the loaded data to the suitable data processor of the model for further processing.
\begin{figure}[ht]
	\centering
    \caption{The metrics summary and a part of automatic visualizations processed by metric visualizer in \our. The experimental dataset is \texttt{ARTS-Laptop14}, an adversarial dataset for ASC.}
	\includegraphics[width=\linewidth]{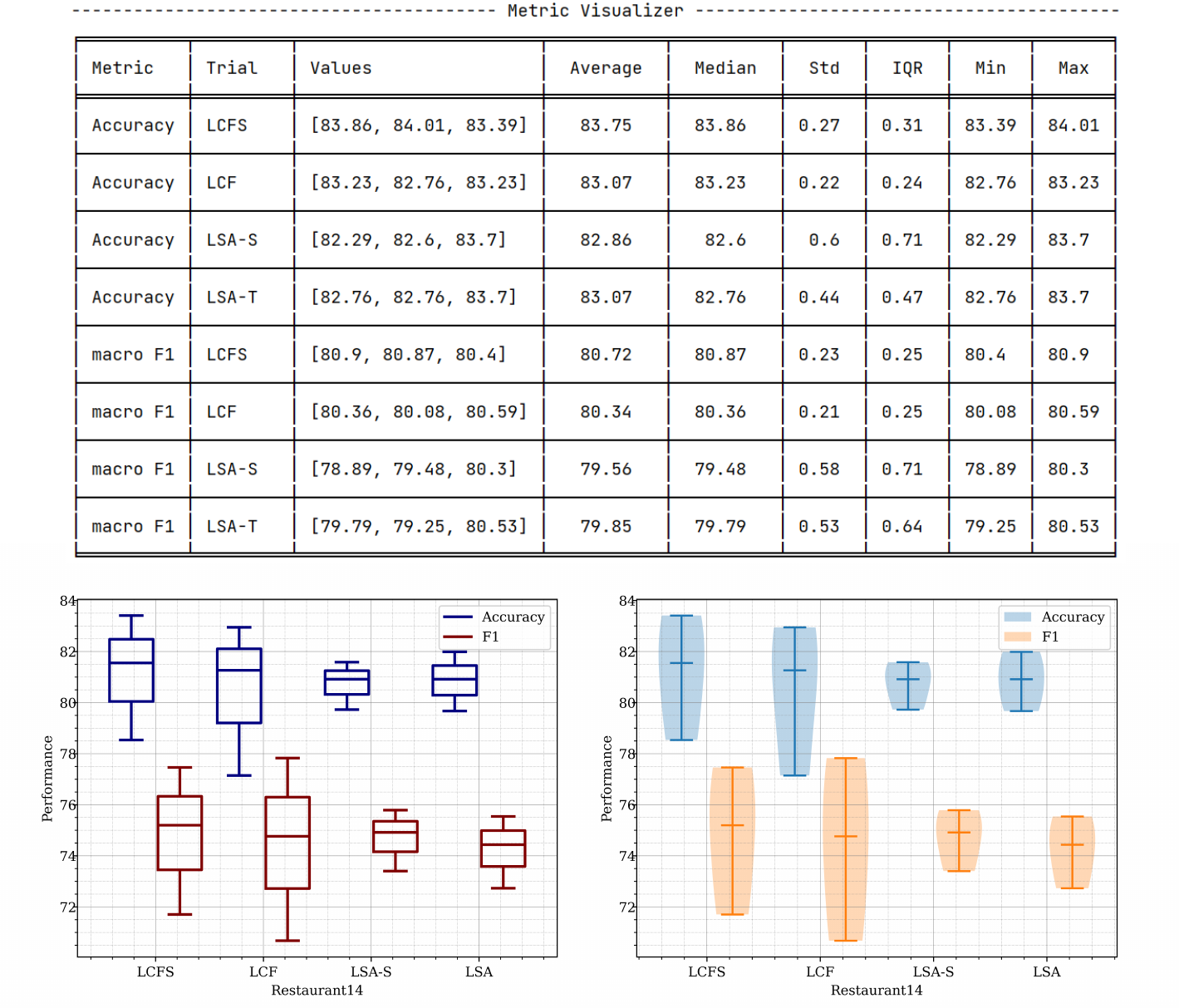}
	\label{fig:mv}
\end{figure}
\subsection{Metric Visualizer}
As a vital effort towards streamlined evaluation and fair comparisons, metric visualizer\footnote{The metric visualizer was developed specifically for \our\ and is available as an independent open-source project at: \url{https://github.com/yangheng95/metric-visualizer}} for \our\ to automatically record, manage, and visualize various metrics (such as Accuracy, F-measure, STD, IQR, etc.). The metric visualizer can track metrics in real-time or load saved metrics records and produce box plots, violin plots, trajectory plots, Scott-Knott test plots, significance test results, etc. An example of auto-generated visualizations is shown in \pref{fig:mv} and more plots and experiment settings can be found in \pref{app:mv}. The metric visualizer streamlines the process of visualizing performance metrics and eliminates potential biases in metric statistics.

\begin{figure*}[ht]
	\centering
    \caption{The community-contributed manual dataset annotation tool provided for \our.}
	\includegraphics[width=\linewidth]{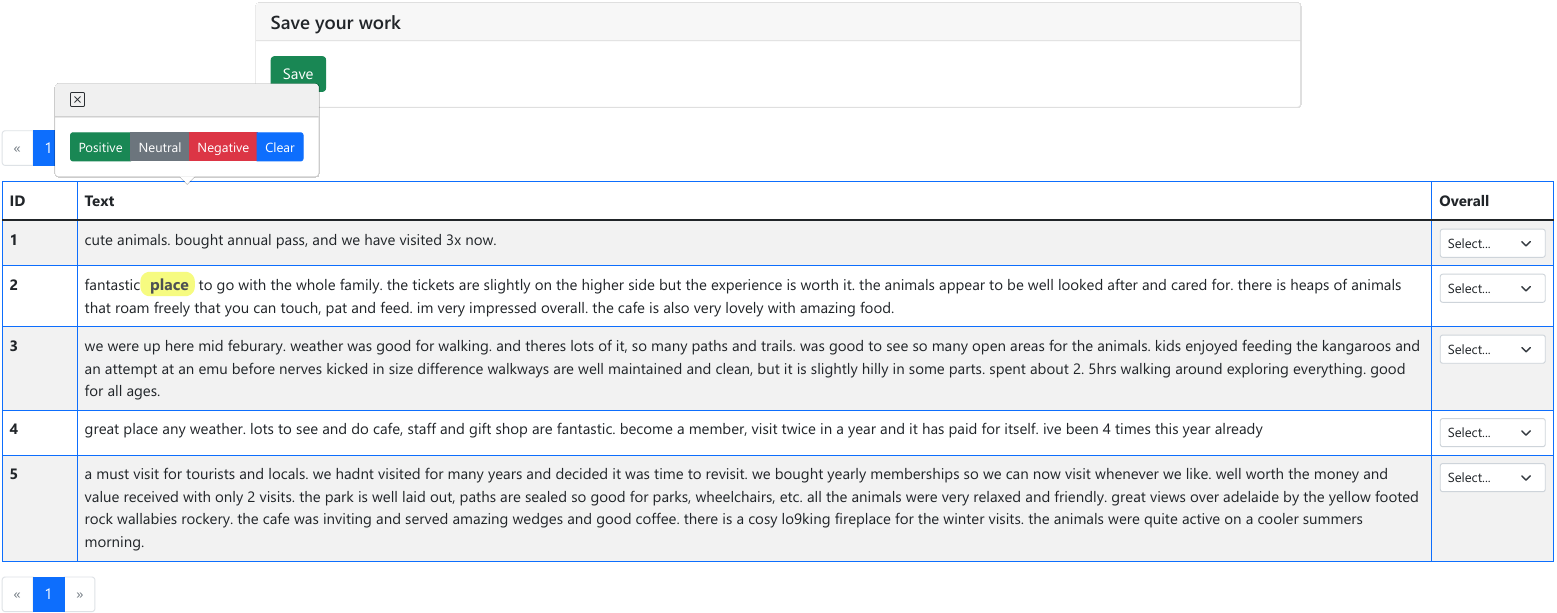}
	\label{fig:dpt}
\end{figure*}

\subsection{Checkpoint Manager}
\begin{figure}[ht]
	\centering
    \caption{A part of available checkpoints for E2E ABSA in \our's model hub.}
	\includegraphics[width=\linewidth]{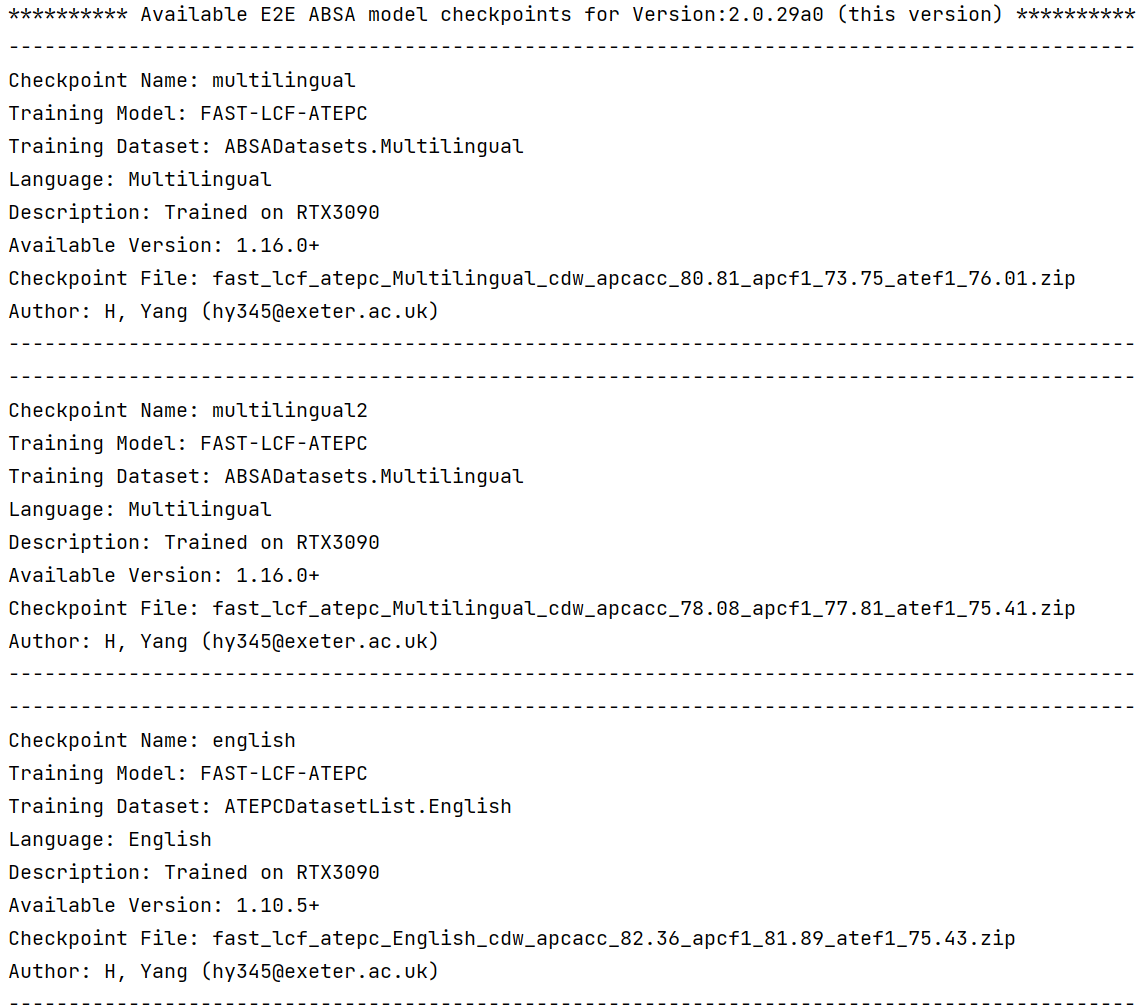}
	\label{fig:ckpts}
\end{figure}

% To make \our\ user-friendly, we have created a model hub, similar to the HuggingFace hub, that allows users to access ABSA inference models with a simple interface. The
Checkpoint manager manages the trained model checkpoints and interacts with the model hub. Users can easily query available checkpoints for different ABSA subtasks and instantiate an inference model by specifying its checkpoint name. Users can query available checkpoints in few lines of code as in Snippet~\ref{snp:ckpt} from the model hub.
The example of available checkpoints is shown in \pref{fig:ckpts}. 

\begin{snippet}[ht]
\caption{The code snippet of available checkpoints.}
\label{snp:ckpt}
\begin{lstlisting}[language=Python,basicstyle=\ttfamily\tiny,frame=shadowbox,keywordstyle=\color{red},breaklines=True]
from pyabsa import available_checkpoints
from pyabsa import TaskCodeOption

checkpoint_map = available_checkpoints(
    # the code of ASC
    TaskCodeOption.Aspect_Polarity_Classification, 
    show_ckpts=True
)

\end{lstlisting}
\end{snippet}

While connecting to the model hub is the most convenient way to get an inference model, we also provide two alternative ways:
\begin{itemize}[leftmargin=*,noitemsep,nolistsep]
\item Searching for trained or cached checkpoints using keywords or paths through the checkpoint manager.
\item Building inference models using trained models returned by the trainers, which eliminates the need for saving checkpoints to disk.
\end{itemize}
The checkpoint manager for any subtask is compatible with GloVe and pre-trained models based on transformers, and with the help of \our's interface, launching an ATESC service requires just a few lines of code.

\section{Featured Functionalities}

\subsection{Data Augmentation}
In ABSA, data scarity can lead to inconsistencies in performance evaluation and difficulties with generalizing across domains. To address this issue, \our\ has adopted an automatic text augmentation method, i.e., \texttt{BoostAug}. This method balances diversity and skewness in the distribution of augmented data. In our experiments, the text augmentation method significantly boosted the classification accuracy and F1 scores of all datasets and models, whereas previous text augmentation techniques had a negative impact on model performance. We refer a comprehensive overview of this text augmentation method to \citet{YangL2022}.

\subsection{Dataset Annotation}
\label{sec:annotation}
Annotating ABSA datasets is more difficult compared to pure text classification. As there is no open-source tool available for annotating ABSA datasets, creating custom datasets becomes a critical challenge. In \our, we have got users rescued by provide a manual annotation interface contributed by the community (referred to as \pref{fig:dpt}), along with an automatic annotation interface.

\paragraph{Manual Annotation}
To ensure accurate manual annotation, our contributor developed a specialized ASC annotation tool\footnote{\href{https://github.com/yangheng95/ABSADatasets/tree/v1.2/DPT}{https://github.com/yangheng95/ABSADatasets/DPT}} for \our. This tool runs on web browsers, making it easy for anyone to create their own datasets with just a web browser. The annotation tool outputs datasets for various ABSA sub-tasks, such as ASC and ATESC sub-tasks, and we even provide an interface to help users convert datasets between different sub-tasks. Check out the community-contributed manual dataset annotation tool in \pref{fig:dpt}

\paragraph{Automatic Annotation}
To make manual annotation easier and address the issue of limited data, we offer an automatic annotation method in \our. This interface is powered by a trained E2EABSA model and uses a hub-powered inference model to extract aspect terms and sentiment polarities. It enables users to quickly expand small datasets with annotated ABSA instances. Check out the following example for a demonstration of the automatic annotation interface:
\begin{snippet}
\caption{The code snippet of automatic annotation.}
\begin{lstlisting}[language=Python,basicstyle=\ttfamily\tiny,frame=shadowbox,keywordstyle=\color{red},breaklines=True]
from pyabsa import make_ABSA_dataset

# annotate "raw_data" using "multilingual" ATESC model 
make_ABSA_dataset(dataset_name_or_path='raw_data', checkpoint='multilingual')
\end{lstlisting}
\end{snippet}

\paragraph{Ensemble Training}
In deep learning, model ensemble is a crucial technique, and it is common to enhance ABSA performance in real-world projects through model ensemble. To simplify the process for users, \our\ provides easy-to-use model ensemble without any code changes. Furthermore, \our\ offers convenient ensemble methods for users to effortlessly augment their training data using built-in datasets from the data center. For example, when \our\ recognizes a model or dataset as a list, it will automatically perform ensemble. We showcase this simple ensemble method in Snippet~\ref{snp:ensemble}.

\begin{snippet}[ht]
\caption{The code snippet of an model ensemble in \our.}
\label{snp:ensemble}
\begin{lstlisting}[language=Python,basicstyle=\ttfamily\tiny,frame=shadowbox,keywordstyle=\color{red},breaklines=True]
import random
from pyabsa import (
    AspectSentimentClassification as ASC,
    ModelSaveOption,
    DeviceTypeOption
)

models = [
        ASC.ASCModelList.FAST_LSA_T_V2,
        ASC.ASCModelList.FAST_LSA_S_V2,
        ASC.ASCModelList.BERT_SPC_V2,
    ]

datasets = [
    ASC.ASCDatasetList.Laptop14,
    ASC.ASCDatasetList.Restaurant14,
    ASC.ASCDatasetList.Restaurant15,
    ASC.ASCDatasetList.Restaurant16,
    ASC.ASCDatasetList.MAMS
]

config = ASC.ASCConfigManager.get_apc_config_english()
config.model = models
config.pretrained_bert = 'roberta-base'
config.seed = [random.randint(0, 10000) for _ in range(3)]

trainer = ASC.ASCTrainer(
    config=config,
    dataset=datasets,
    checkpoint_save_mode=ModelSaveOption.SAVE_MODEL_STATE_DICT,
    auto_device=DeviceTypeOption.AUTO,
)
trainer.load_trained_model()

\end{lstlisting}
\end{snippet}

\paragraph{Ensemble Inference}

\our\ includes an ensemble inference module for all subtasks, which enables users to aggregate the results of multiple models to produce a final prediction, thereby leveraging the strengths of each individual model and resulting in improved performance and robustness compared to using a single model alone. We provide an example of ensemble inference in Snippet~\ref{snp:ensemble2}.

\begin{snippet}[ht]
\caption{The code snippet of an model ensemble in \our.}
\label{snp:ensemble2}
\begin{lstlisting}[language=Python,basicstyle=\ttfamily\tiny,frame=shadowbox,keywordstyle=\color{red},breaklines=True]
from pyabsa.utils import VoteEnsemblePredictor

checkpoints = {
    ckpt: APC.SentimentClassifier(checkpoint=ckpt)
    # use the findfile module to search all available checkpoints 
    for ckpt in findfile.find_cwd_dirs(or_key=["laptop14"])
}

ensemble_predictor = VoteEnsemblePredictor(
    checkpoints, weights=None, numeric_agg="mean", str_agg="max_vote"
)

ensemble_predictor.predict("The [B-ASP]food[E-ASP] was good!")

\end{lstlisting}
\end{snippet}

\section{Conclusions and Future Work}

We present \our, a modularized framework for reproducible ABSA. Our goal was to democratize the reproduction of ABSA models with a few lines of code and provide an opportunity of implementing ideas with minimal modifications on our prototypes. Additionally, the framework comes equipped with powerful data augmentation and annotation features, largely addressing the data scarity of ABSA. In the future, we plan to expand the framework to include other ABSA subtasks, such as aspect sentiment triplet extraction.

\section*{Acknowledgements}
We appreciate all contributors who help \our\, e.g., committing code or datasets; the community's support makes \our\ even better. Furthermore, we appreciate all ABSA researchers for their open-source models that improve ABSA.

\bibliography{anthology,custom}
\bibliographystyle{acl_natbib}

\appendix
\section{Related Works}

% In recent years, a large number of outstanding open-source ASC \cite{LiCFMWH20,TianCS21,LiZZZW21,WangSLZC21} and ATESC \cite{LiBLLY18,XuLSY18,MaLWXW19,Yang2019,YangLQSS20} models have been proposed. However, the related open-source repositories for these models usually lack inference support, and most of them are no longer maintained. There are two works most similar to \our, ABSA-PyTorch and Aspect-based Sentiment Analysis, respectively. ABSA-PyTorch \cite{SongWJLR19} incorporates multiple reimplemented third-party GloVe-based and BERT-based models as an early effort to propagate fair comparisons of accuracy and F1 amongst models. Nevertheless, ABSA-PyTorch is no longer maintained and only supports the ASC subtask. ASC subtasks are also handled by Aspect-based Sentiment Analysis \cite{Scalac2020}, which provides an ASC inference interface based on constrained models. \our\ is a research- and application-friendly framework that supports a number of ABSA subtasks and includes multilingual, open-source ABSA datasets. We developed instant inference interfaces for ASC and ATESC subtasks, which facilitate the implementation of multilingual ABSA services, using inspiration from Transformers.

In recent years, many open-source models have been developed for aspect-based sentiment classification (ASC)~\cite{LiCFMWH20,TianCS21,LiZZZW21,WangSLZC21} and aspect term extraction and sentiment classification (ATESC)~\cite{LiBLLY18,XuLSY18,MaLWXW19,Yang2019,YangLQSS20}. However, the open-source repositories for these models often lack the capability to make predictions, and many are no longer being maintained. Two works similar to \our\ are ABSA-PyTorch~\cite{SongWJLR19} and Aspect-based Sentiment Analysis. \texttt{ABSA-PyTorch} combined multiple third-party models to facilitate fair comparisons of accuracy and F1, but it is now outdated and only supports the ASC task. \texttt{Aspect-based Sentiment Analysis}~\cite{Scalac2020} also handles ASC, but with limited models. \our\ is a research-friendly framework that supports multiple aspect-based sentiment analysis (ABSA) subtasks and includes multilingual, open-source ABSA datasets. The framework has instant inference interfaces for both aspect-based sentiment classification (ASC) and aspect-term extraction and sentiment classification (ATESC) subtasks, facilitating the implementation of multilingual ABSA services. \our\ sets itself apart from other similar works, such as ABSA-PyTorch and Aspect-based Sentiment Analysis, by being actively maintained and supporting multiple ABSA subtasks.

\section{Model Evaluation}

\label{app:eval}
We present the experimental results of various models on different datasets, which may help users choose a suitable model for their projects.

\begin{sidewaystable}[hbtp]
    \centering
    \caption{The evaluation of the performance of the ASC and ATESC models on the datasets available in \our. The results in parentheses are the standard deviations. These results were obtained through 10 epochs of training using the default settings. The multi-language dataset includes all the built-in datasets from \our. The absence of results for some datasets using syntax-based models is indicated by "--".}
    \resizebox{\linewidth}{!}{
    \begin{tabular}{|c|c|c|c|c|c|c|c|c|c|c|c|c|c|c|c|c|c||c|c|}
    \hline
    \multirow{1}[4]{*}{\textbf{ASC Model}} & \multirow{1}[4]{*}{\textbf{Task}} & \multicolumn{2}{c|}{\textbf{English}} & \multicolumn{2}{c|}{\textbf{Chinese}} & \multicolumn{2}{c|}{\textbf{Arabic}} & \multicolumn{2}{c|}{\textbf{Dutch}} & \multicolumn{2}{c|}{\textbf{Spanish}} & \multicolumn{2}{c|}{\textbf{French}} & \multicolumn{2}{c|}{\textbf{Turkish}} & \multicolumn{2}{c||}{\textbf{Russian}} & \multicolumn{2}{c|}{\textbf{Multilingual}} \\
\cline{3-20}          &       & Acc$_{ASC}$ &  F$1_{ASC}$ & Acc$_{ASC}$ &  F$1_{ASC}$ & Acc$_{ASC}$ &  F$1_{ASC}$ & Acc$_{ASC}$ &  F$1_{ASC}$  & Acc$_{ASC}$ &  F$1_{ASC}$ & Acc$_{ASC}$ &  F$1_{ASC}$  & Acc$_{ASC}$ &  F$1_{ASC}$ & Acc$_{ASC}$ &  F$1_{ASC}$  & Acc$_{ASC}$ &  F$1_{ASC}$ \\
    \hline
    \hline
    % \texttt{BERT-BASE-ASC} & \multirow{7}[24]{*}{} &      & -- &       & -- &       & -- &       & -- &       & -- &       & --  &       &  --  &       & -- &       & -- \\
    \texttt{BERT-SPC} & \multirow{5}[24]{*}{\begin{sideways}ASC\end{sideways}}
    & 84.57(0.44) & 81.98(0.22) & 95.27(0.29) & 94.10(0.40) & 95.27(0.29) & 94.10(0.40) & 89.12(0.21) & 74.64(0.68) & 86.29(0.51) & 68.64(0.26) & 86.14(0.63) & 75.59(0.45) & 85.27(0.34) & 65.58(0.07) & 87.62(77.13) & 77.13(0.08) & 87.19(0.36) & 80.93(0.21) \\
    \cline{1-1}\cline{3-20}    \texttt{DLCF-DCA} &       & 84.05(0.06) & 81.03(1.05) & 95.69(0.22) & 94.74(0.30) & 89.23(0.15) & 75.68(0.01) & 86.93(0.38) & 72.81(1.57) & 86.42(0.49) & 76.29(0.10) & 90.42(0.0) & 73.69(0.82) & 85.96(1.71) & 67.59(1.61) & 87.00(0.41) & 74.88(0.41) & 87.80(0.01) & 81.62(0.20) \\
    % \cline{1-1}\cline{3-20}    \texttt{DLCFS-DCA} &       & 84.12(0.12) & 81.34(0.01) &  & -- &       & -- &       & -- &       & -- &       & -- &       & -- &       & -- &       & -- \\
    \cline{1-1}\cline{3-20}    \texttt{Fast-LCF-ASC} &       & 84.70(0.05) & 82.00(0.08) & 95.98(0.02) & 95.01(0.05) & 89.82(0.06) & 77.68(0.33) & 86.42(0.13) & 71.36(0.53) & 86.35(0.28) & 75.10(0.14) & 91.59(0.21) & 72.31(0.26) & 86.64(1.71) & 67.00(1.63) & 87.77(0.15) & 74.17(0.47) &  87.66(0.15) & 81.34(0.25) \\
    \cline{1-1}\cline{3-20}    \texttt{Fast-LCFS-ASC} &       & 84.27(0.09) & 81.60(0.17) & 95.67(0.32) & 94.40(0.55) & -- & -- & -- & -- & -- & -- & -- & -- &   -- & -- & --& -- & -- & -- \\
    % \cline{1-1}\cline{3-20}    \texttt{LCA-BERT} &       &       & -- &       & -- &       & -- &       & -- &       & -- &       & -- &       & -- &       & -- &       &  --\\
    
    \cline{1-1}\cline{3-20}    \texttt{LCF-BERT} &       & 84.81(0.29) & 82.06(0.06) & 96.30(0.05) & 95.45(0.05) & 89.80(0.13) & 77.60(0.44) & 86.55(0.76) & 70.67(0.41) & 85.52(0.42) & 74.03(0.75) & 91.86(0.21) & 75.26(0.37) & 89.73(0.05) & 68.57(1.09) & 87.41(0.21) & 74.71(0.17)  & 87.86(0.09) & 82.01(0.46) \\
    
    \cline{1-1}\cline{3-20}    \texttt{LCFS-BERT} &       & 84.49(0.13) & 81.46(0.05) & 95.32(0.39) & 94.23(0.56) & 88.89(0.11) & 75.41(0.37) & 87.94(0.43) & 72.69(1.01) & 84.61(0.21) & 71.98(1.25) & 90.83(0.41) & 73.87(1.45) & 88.36(0.68) & 69.21(0.86) & 87.15(0.15) & 74.99(0.44) & 87.55(0.22) & 81.58(0.13) \\
    
    \cline{1-1}\cline{3-20}    \texttt{Fast-LSA-T} &       & 84.60(0.29) & 81.77(0.44) & 96.05(0.05) & 95.10(0.05) & 89.25(0.38) & 77.25(0.43) & 86.04(0.0) & 70.02(0.75) & 86.07(0.14) & 73.52(0.53) & 91.93(0.27) & 74.21(0.60) & 88.01(1.03) & 66.74(0.61) & 88.24(0.10) & 76.91(1.10) & 87.56(0.13) & 81.01(0.56) \\
    \cline{1-1}\cline{3-20}    \texttt{Fast-LSA-S} &       & 84.15(0.15) & 81.53(0.03) & 89.55(0.11) & 75.87(0.25) & -- & -- & -- & -- & -- & -- & -- & -- & -- & -- & -- & -- &  -- &  --\\
    \cline{1-1}\cline{3-20}    \texttt{Fast-LSA-P} &       & 84.21(0.06) & 81.60(0.23) & 95.27(0.29) & 94.10(0.40) & 89.12(0.21) & 74.64(0.68) & 86.29(0.51) & 68.64(0.26) & 86.14(0.63) & 75.59(0.45) & 90.77(0.07) & 74.60(0.82) & 85.27(0.34) & 65.58(0.07) & 87.62(0.06) & 77.13(0.08) & 87.81(0.24) & 81.58(0.56) \\
    \hline
    \hline
    \multirow{1}[4]{*}{\textbf{ATESC Model}} & \multirow{1}[4]{*}{\textbf{Task}} & \multicolumn{2}{c|}{\textbf{English}} & \multicolumn{2}{c|}{\textbf{Chinese}} & \multicolumn{2}{c|}{\textbf{Arabic}} & \multicolumn{2}{c|}{\textbf{Dutch}} & \multicolumn{2}{c|}{\textbf{Spanish}} & \multicolumn{2}{c|}{\textbf{French}} & \multicolumn{2}{c|}{\textbf{Turkish}} & \multicolumn{2}{c||}{\textbf{Russian}} & \multicolumn{2}{c|}{\textbf{Multilingual}} \\
\cline{3-20}          &       &  F$1_{ASC}$ & F1$_{ATE}$ &  F$1_{ASC}$ & F1$_{ATE}$ &  F$1_{ASC}$ & F1$_{ATE}$ &  F$1_{ASC}$ & F1$_{ATE}$  &  F$1_{ASC}$ & F1$_{ATE}$ &  F$1_{ASC}$ & F1$_{ATE}$  &  F$1_{ASC}$ & F1$_{ATE}$ &  F$1_{ASC}$ & F1$_{ATE}$  &  F$1_{ASC}$ & F1$_{ATE}$ \\
    \hline
    \hline
    \texttt{BERT-ATESC} & \multirow{3}[10]{*}{\begin{sideways}ATESC\end{sideways}} & 72.70(0.48) & 81.66(1.16) &  94.12(0.12) & 64.86(0.50) & 71.18(0.34) & 71.18(0.34) & 71.06(1.09) & 78.31(1.40) & 69.29(0.07) & 83.54(0.59) & 69.26(0.07) & 83.54(0.59) & 67.28(0.37) & 72.88(0.37) & 70.46(0.54) & 77.64(0.19) & 75.13(0.15) & 80.9(0.02) \\
    \cline{1-1}\cline{3-20}    \texttt{Fast-LCF-ASESC} &  & 79.23(0.07) & 81.78(0.12) & 94.32(0.29) & 84.15(0.39) & 67.38(0.11) & 70.30(0.41) & 73.67(0.89) & 78.59(0.69) & 71.24(0.47) & 83.37(1.21) & 71.24(0.47) & 82.06(0.67) & 67.64(1.28) & 73.46(0.87) & 71.28(0.37) & 76.90(0.52) & 78.96(0.13) & 80.31(0.19) \\
    \cline{1-1}\cline{3-20}    \texttt{Fast-LCFS-ASESC} &  & 75.82(0.03) & 81.40(0.59) & 93.68(0.25) & 84.48(0.32) & -- & -- & -- & -- & -- & -- & -- & -- & --   & -- & -- & -- & -- & -- \\
    \cline{1-1}\cline{3-20}    \texttt{LCF-ATESC} &       & 77.91(0.41) & 82.34(0.35) & 94.00(0.38) & 84.64(0.38) & 67.30(0.17) & 70.52(0.28) & 71.85(1.53) & 79.94(1.70) & 70.19(0.24) & 84.22(0.83) & 70.76(0.92) & 82.16(0.38) & 69.52(0.54) & 74.88(0.08) & 71.96(0.28) & 79.06(0.52) & 80.63(0.35) & 80.15(1.18) \\
    \cline{1-1}\cline{3-20}    \texttt{LCFS-ATESC} &      & 75.85(0.22) & 85.00(1.58) & 93.52(0.09) & 84.92(0.11) & -- & -- & -- & -- & -- & -- & -- & -- & --   & -- & -- & -- & -- & --  \\
    \hline
    \end{tabular}%
}
\label{app:performance}
\end{sidewaystable}

\section{Metric Visualization in \our}
\label{app:mv}
\subsection{Code for Auto-metric Visualization}

\our\ provides standardised methods for monitoring metrics and metric visualisations. PyASBA will automatically generate trajectory plot, box plot, violin plot, and bar charts based on metrics to evaluate the performance differences across models, etc. This example aims at evaluating the influence of maximum modelling length as a hyperparameter on the performance of the \texttt{FAST-LSA-T-V2} model on the Laptop14 dataset.

\begin{lstlisting}[language=Python,basicstyle=\ttfamily\tiny,frame=shadowbox,keywordstyle=\color{red},breaklines=True]
import random
import os
from metric_visualizer import MetricVisualizer

from pyabsa import AspectSentimentClassification as ASC

config = ASC.ASCConfigManager.get_config_english()
config.model = ASC.ASCModelList.FAST_LSA_T_V2
config.lcf = 'cdw'

# each trial repeats with different seed
config.seed = [random.randint(0, 10000) for _ in range(3)]

MV = MetricVisualizer()
config.MV = MV

max_seq_lens = [50, 60, 70, 80, 90]

for max_seq_len in max_seq_lens:
    config.max_seq_len = max_seq_len
    dataset = ABSADatasetList.Laptop14
    Trainer(config=config,
            dataset=dataset, 
            auto_device=True  
            )
    config.MV.next_trial()

save_prefix = os.getcwd()
# save fig into .tex and .pdf format
MV.summary(save_path=save_prefix)

MV.to_execl(save_path=os.getcwd() + "/example.xlsx")  # save to excel
MV.box_plot(no_overlap=True, save_path="box_plot.png")
MV.violin_plot(no_overlap=True, save_path="violin_plot.png")
MV.scatter_plot(save_path="scatter_plot.png")
MV.trajectory_plot(save_path="trajectory_plot.png")

MV.scott_knott_plot()
MV.A12_bar_plot()
\end{lstlisting}

\subsection{Automatic Metric Visualizations}
There are some visualization examples auto-generated by \our. Note that the metrics are not stable on small datasets. 

\begin{figure}[ht]
	\centering
	\includegraphics[width=\linewidth]{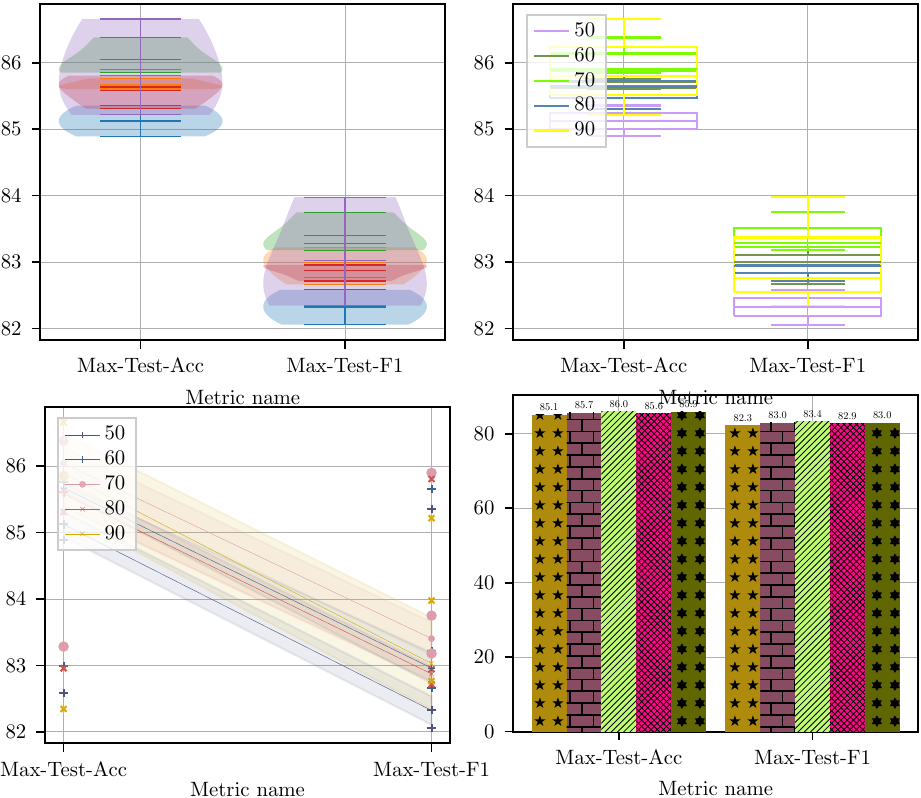}
	\caption{An example of automated -metric visualizations of the \texttt{Fast-LSA-T-V2} model grouped by metric names.}
\end{figure}

\begin{figure}[ht]
	\centering
	\includegraphics[width=\linewidth]{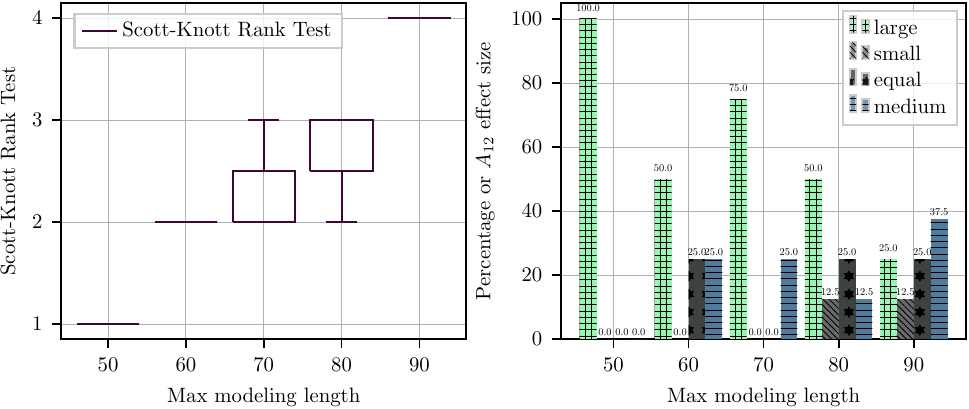}
	\caption{The significance level visualizations of the \texttt{Fast-LSA-T-V2} grouped by different max modeling length. The left is scott-knott rank test plot, while the right is A12 effect size plot.}
	\label{fig:plot}
\end{figure}

\end{document}